\documentclass[lettersize, 10 pt, journal, twoside]{IEEEtran}

\IEEEoverridecommandlockouts                              

\usepackage{amsmath} 
\usepackage{amssymb}  
\usepackage{bm, balance}
\usepackage{algorithm,algpseudocode}
\usepackage[small]{caption}
\usepackage{multicol}
\usepackage{subcaption}
\usepackage{graphicx}
\usepackage{xcolor,mathrsfs,array}
\usepackage[export]{adjustbox}
\usepackage[utf8]{inputenc}
\usepackage[T1]{fontenc}
\usepackage{cite}
\usepackage{flushend}

\usepackage[linkbordercolor={1 1 1},citebordercolor={1 1
  1},urlbordercolor={0.0 0.0
  0.0},urlcolor=blue,colorlinks=true,linkcolor=black,citecolor=black]{hyperref}
\usepackage{url}

\DeclareUnicodeCharacter{2217}{*}

\usepackage{shortcuts} 

\begin{document}

\title{
Natural Multicontact Walking for Robotic Assistive Devices via Musculoskeletal Models and Hybrid Zero Dynamics}

\author{Kejun Li$^{1}$, \textit{Graduate Student Member, IEEE,} Maegan Tucker$^{2}$, \textit{Graduate Student Member, IEEE,} \\ Rachel Gehlhar$^{2}$, Yisong Yue$^{3}$, and Aaron D. Ames$^{2,3}$, \textit{Fellow, IEEE} 
\thanks{Manuscript received September 9, 2021; accepted January 24, 2022. Date of publication; date of current version. This letter was recommended for publication by Associate Editor Andrea Del Prete and Editor Abderrahmane Kheddar upon evaluation of the reviewers' comments. This work was supported in part by NSF GRF under Grant DGE‐1745301, in part by Wandercraft, and in part by the ZEITLIN Fund, and conducted under IRB No. 21-0693.}
\thanks{Kejun Li is with the Department of Biology and Biological Engineering, California Institute of Technology, Pasadena, CA 91125 USA (email: kli5@caltech.edu).}
\thanks{Maegan Tucker, Rachel Gehlhar, and Aaron D. Ames are with the Department of Mechanical and Civil Engineering, California Institute of Technology, Pasadena, CA 91125 USA (email: mtucker@caltech.edu; rgehlhar@caltech.edu; ames@caltech.edu).}%
\thanks{Yisong Yue and Aaron D. Ames are with the Department of Computing and Mathematical Sciences, California Institute of Technology, Pasadena, CA 91125 USA (email: yyue@caltech.edu; ames@caltech.edu).}
\thanks{This letter has supplementary downloadable material available at
https://doi.org/10.1109/LRA.2022.3149568, provided by the authors.}
\thanks{Digital Object Identifier (DOI): see top of this page}
}

\markboth{IEEE Robotics and Automation Letters. Preprint Version. Accepted 1, 2022}
{Li \MakeLowercase{\textit{et al.}}: Natural Multicontact Walking via Musculoskeletal Models and Hybrid Zero Dynamics}

\maketitle

\begin{abstract}
Generating stable walking gaits that yield natural locomotion when executed on robotic-assistive devices is a challenging task that often requires hand-tuning by domain experts. This paper presents an alternative methodology, where we propose the addition of musculoskeletal models directly into the gait generation process to intuitively shape the resulting behavior. In particular, we construct a multi-domain hybrid system model that combines the system dynamics with muscle models to represent natural multicontact walking. \new{Provably} stable walking gaits can then be generated for this model via the hybrid zero dynamics (HZD) method. We experimentally apply our integrated framework towards achieving multicontact locomotion on a dual-actuated transfemoral prosthesis, AMPRO3, \new{for two subjects}. The results demonstrate that enforcing \new{muscle model constraints} produces gaits that yield natural locomotion \new{(as analyzed via comparison to motion capture data and electromyography). Moreover, gaits generated with our framework were strongly preferred by the non-disabled prosthetic users as compared to gaits generated with the nominal HZD method, even with the use of systematic tuning methods.} We conclude that the novel approach of combining robotic walking methods (specifically HZD) with muscle models successfully generates anthropomorphic robotic-assisted locomotion.
\end{abstract}


\IEEEpeerreviewmaketitle

\section{Introduction}
\label{sec: intro}
\IEEEPARstart{W}{hile} there is extensive literature on the biomechanics surrounding non-disabled human walking \cite{winter2009biomechanics}, it is poorly understood how to translate this natural and efficient bipedal locomotion to robotic platforms, especially in the context of robotic assistive devices which necessitate cooperation with human users. In this work, we aim to achieve stable and natural assisted locomotion by incorporating musculoskeletal models directly into a multi-domain gait generation process. 

Even though bipedal locomotion is seemingly effortless for humans, achieving stable walking on robotic platforms is challenging as it requires accounting for discrete impact events, underactuation, and complicated nonlinear dynamics.
Existing methods that have successfully demonstrated stable robotic locomotion include reduced-order models \cite{kajita2002realtime,erbatur2008natural,hereid2014dynamic,rezazadeh2015toward}, model-based gait generation \cite{mombaur2009using,powell2015model,reher2021dynamic}, and reinforcement learning \cite{benbrahim1997biped,kohl2004policy,siekmann2021sim,duan2021learning,xie2018feedback}. Of these approaches, we are particularly interested in the \emph{Hybrid Zero Dynamics} (HZD) method \cite{grizzle2001asymptotically,westervelt2003hybrid}, which is a mathematical approach leveraging the hybrid system model of locomotion, and capable of synthesizing provably stable \cite{ames2014rapidly} dynamic walking gaits on bipedal robots, encoded by impact-invariant periodic orbits. This method has been demonstrated on a number of robotic platforms and behaviors, including dynamic multicontact walking on 3D robots \cite{reher2021dynamic}. 

While the HZD method yields provably stable walking, one drawback of the approach is the reliance on a carefully-tuned optimization problem to obtain converged solutions for stable period orbits. \new{This tuning process often entails domain experts modifying the cost and constraints until ``good'' gaits are found. Obtaining a satisfactory gait is especially challenging for robotic assistive devices since they require not only stable but also natural locomotion to alleviate high energy expenditure during ambulation, which demands even more tuning to sufficiently constrain the problem.}

\begin{figure}[tb]
         \centering \includegraphics[width=\linewidth]{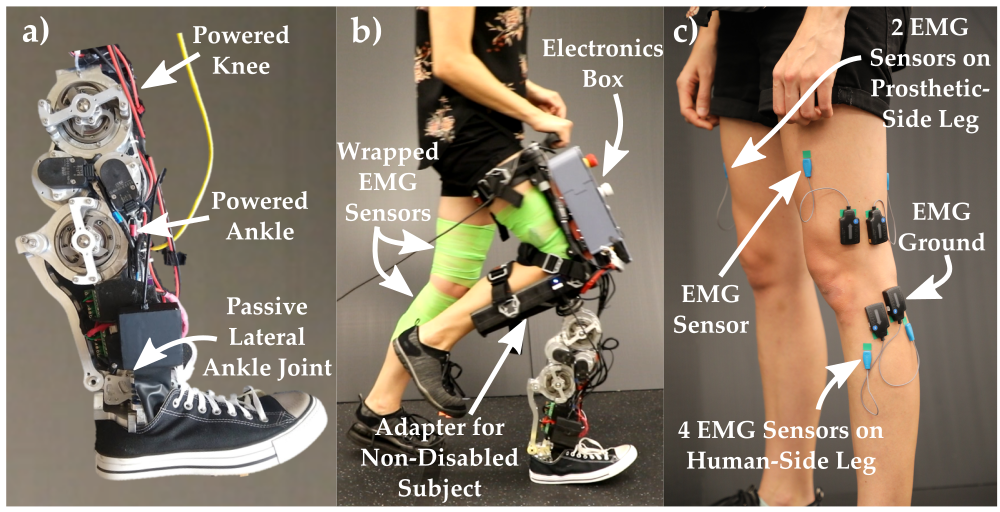}
         \caption{a) AMPRO3 prosthesis, b) Non-disabled subject wearing the device during multicontact locomotion, c) placement of the surface mount electrodes for electromyography (EMG).}
          \label{fig:exsetup}
\end{figure}

\new{Other approaches towards realizing natural walking include modifying the HZD method to obtain gaits that resemble walking recorded by motion capture \cite{ames2014human, zhao2017multi} and optimizing joint-level trajectories for experimental metrics such as electromyography (EMG) signals and metabolic expenditure \cite{zhang2017human,han2021selection}. While these methods yield natural behavior, they are data-driven and thereby heavily rely on the quality and quantity of the data. Moreover, such data is behavior-specific and not always accessible. A separate approach that does not rely on gait generation is to directly control the walking based on real-time EMG feedback \cite{au2005emg,hoover2011configuration,wu2011electromyography,wang2013proportional,cimolato2020hybrid}. While this methodology also successfully yields natural locomotion, it has no theoretical guarantees of stability and relies on careful tuning of the musculoskeletal model.}



In this work, we present an alternative approach based upon hybrid system models of locomotion that utilize musculoskeletal models --- to our best knowledge, this is the first time these two modeling paradigms have been combined. Our proposed integrated framework both enjoys the theoretical guarantees of stability via the HZD method, while also achieving natural locomotion via the musculoskeletal models. Since humans usually self-select gaits that are physiologically and mechanically energy efficient \cite{waters1999energy}, we hypothesize that generating stable gaits that satisfy muscle model constraints would naturally lead to more anthropomorphic and efficient behavior that respects physiological limits.


This hypothesis will be evaluated throughout this paper by first generating multicontact walking gaits utilizing the HZD method coupled with musculoskeletal models, followed by the experimental implementation on a dual-actuated transfemoral prosthesis, AMPRO3, shown in Fig. \ref{fig:exsetup}. The experimental results demonstrate that the novel combination of muscluloskeletal models with HZD results in natural multicontact locomotion, as \new{quantified by comparisons with motion capture data and via electromyography (EMG).} 

\section{Muscle Model}
\label{sec: musclemodel}
In this section, we introduce how a single muscle-tendon unit (MTU) is modeled. Later, in Sec. \ref{sec: framework}, we will provide details on how we extend these muscle models to multiple muscles and incorporate them into the Hybrid Zero Dynamics (HZD) gait generation framework. 

\begin{figure}[tb]
         \centering \includegraphics[width=\linewidth]{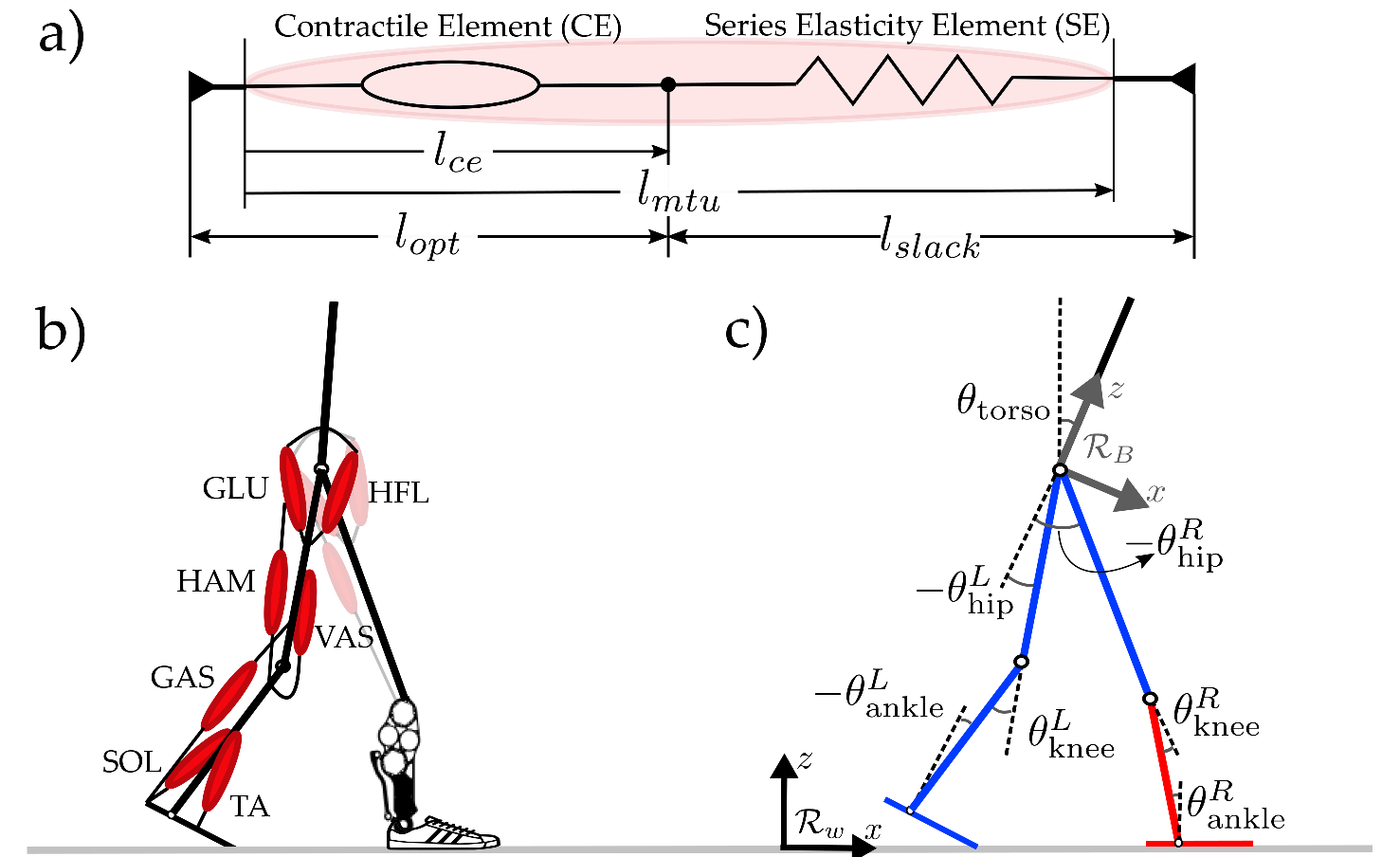}
         \caption{\new{a) A single muscle tendon unit (MTU) consists of a contractile element (CE) and a series elasticity element (SE). The length of CE and SE is denoted by $l_{ce}$ and $l_{se}$. At the reference angle ($\theta_{ref}$), these lengths are equal to $l_{ce} = l_{opt}$ and $l_{se} = l_{slack}$. b) Human-prosthesis system with the following seven labeled muscles on the intact leg: gluteus (GLU), hamstrings (HAM), gastrocnemius (GAS), soleus (SOL), hip flexors (HFL), and vastus (VAS), and tibialis anterior (TA). Three muscles (GLU, HAM, HFL) are also considered on the prosthetic leg side. c) Illustration of system coordinates, including the base and world frames.}}
          \label{fig:model}
\end{figure}

\newsubsec{Muscle-tendon Unit (MTU)}
We model each muscle as a two-element Hill-type muscle-tendon unit \new{\cite{geyer2003positive}} with a contractile element (CE) and a series elastic element (SE) as shown in Fig. \ref{fig:model}a. The constant parameters of each muscle are defined in \cite{geyer2010muscle,geyer2003positive}.

\newsubsec{MTU Length}
The length of an individual MTU, denoted by $l_{mtu} \in \R$, is modeled as $l_{mtu} = l_{se} + l_{ce}$, where $l_{ce} \in \R$ is the length of the contractile element (CE), and $l_{se} \in \R$ is the length of the series elasticity element (SE). 
Since the relative change of $l_{mtu}$ depends on the individual joint angle $\theta \in \R$, with the collection of $d$ joint angles denoted $q \in \R^{d}$, in practice we model the MTU length as a function of $q$:
\begin{align}
    l_{mtu}(q) &= l_{opt} + l_{slack} - \sum_{j = 1}^{j_N} \Delta l_{mtu}(\theta_j),
\end{align}
\new{where $l_{opt}, \, l_{slack} \in \R$ are respectively the reference lengths of CE and SE at the reference angle $\theta_{ref} \in \R$. These reference parameters are constants taken from \cite{geyer2010muscle}.} We use $\sum_{j=1}^{j_N} \Delta l_{mtu}(\theta_j)$ to denote the total change in length of the MTU based on the joint angles of each joint spanned by the MTU, out of a total of $j_N \in \{1,2\}$ joints. The joints spanned by each MTU are illustrated in Fig.\ref{fig:model}b. 
The individual change in length due to a single joint, $\Delta l_{mtu}(\theta) \in \R$, is given by:
\begin{align}
   \Delta l_{mtu}(\theta) &= \begin{cases}
     \rho \new{r_0} (\theta - \theta_{ref}), & \text{ for hip}\\
    \rho \new{r_0} \big[\sin(\theta - \theta_{max}) \\
    \quad -\sin(\theta_{ref} - \theta_{max})\big], & \text{otherwise}
    \end{cases} 
    \label{eq: mtu_deltaL}
\end{align}
The constant $\rho \in \R$ is a parameter that ensures the fiber length is within the physiological limits and accounts for muscle pennation angles (the angle between the longitudinal axis of the entire muscle and its fibers that increases as the tension increases in the muscle), \new{and $r_0 \in \R$ is a parameter denoting the constant contribution of the MTU lever-arm.} For the MTUs that span two joints, $\Delta l_{mtu}(\theta)$ is calculated separately with different reference angles $\theta_{ref}$ for each joint.


\begin{figure*}[t!]
    \centering
    \includegraphics[width=\linewidth]{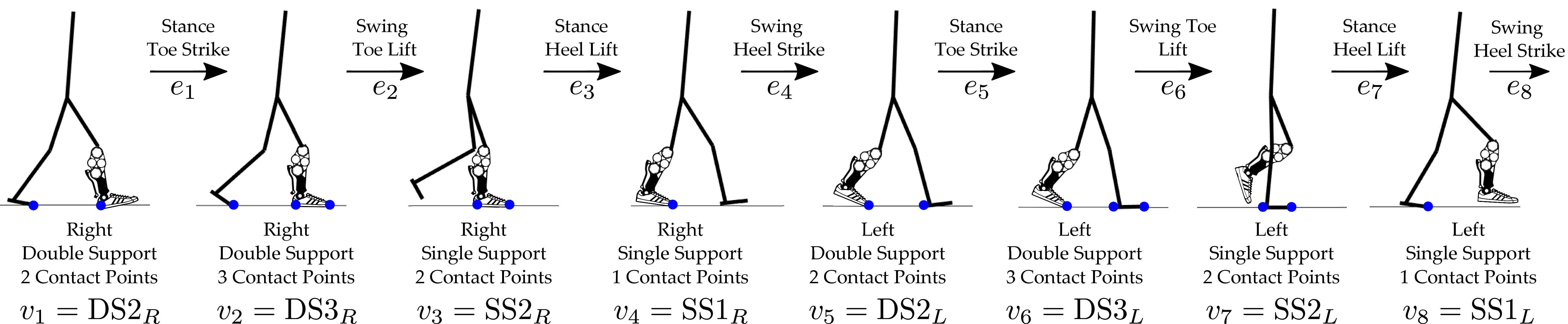}
    \caption{A complete gait cycle from right heel strike to right heel strike. The gait cycle is described using the directed cycle $\Gamma = (V,E)$ with the vertices $V = \{v_1,\dots,v_8\}$ and edges $E = \{e_1,\dots,e_8\}$ illustrated in the figure. The naming convention is based on the stance leg of the step and the number of contact points. If both legs are in contact, the domain is considered as a double support domain.}
    \label{fig:domains}
\end{figure*}

\newsubsec{MTU Force-Length and Force-Velocity Relationships}
The velocity of the CE contraction is denoted by $v_{ce} \in \R$ and is constrained to satisfy the relationship $l_{ce} = \int v_{ce}dt$.
Depending on an MTU's instantaneous value of $l_{ce}$ and $v_{ce}$, the amount of force the MTU is capable of exerting differs. This is described by the following force-length ($f_l$) and force-velocity ($f_v$) relationships:
\begin{align}
 &  f_l(l_{ce}) = \exp \left( \log(c) \left| \frac{l_{ce}-l_{opt}}{l_{opt}w} \right| ^3 \right), \\
  &  f_v(v_{ce}) = \begin{cases}
    \frac{v_{max} - v_{ce}}{v_{max} + Kv_{ce}}, & \text{ if } v_{ce} < 0  \\
    N+\frac{(N-1)(v_{max} + v_{ce})}{7.56Kv_{ce}-v_{max}}, & \text{ if } v_{ce} \geq 0 \end{cases}
\end{align} 
where \new{the residual force factor $c = 0.05$} and $N, \, v_{max}, \, w, \, K \in \R$ are all muscle-dependent constants. Specifically, $N$ is the eccentric force enhancement (modeling the increase in muscle force during active stretch), $v_{max}$ is the maximum contractile velocity, and $w$ and $K$ are parameters that shape the force-length and force-velocity curves, respectively.

Similarly, the MTU force also depends on $l_{se}$. This is modeled using an additional force-length relationship:
\begin{align}
        & f_{se}(l_{se}) = \begin{cases}  \left(\frac{l_{se}-l_{slack}}{l_{slack}(\varepsilon_{ref})}\right)^2, & \text{ if } l_{se} \geq l_{slack}   \\
    0, & \text{ otherwise} 
    \end{cases}
\end{align}
where the $\varepsilon_{ref} \in \R$ is a constant parameter denoting the MTU strain when $f_{se}(l_{se}) = 1$. Note that in the actual implementation, we used a continuous function, fitted via least squares regression, to replace the piece-wise functions for $f_{se}$ and $f_v$ since continuous functions are required for the implementation of a nonlinear optimization program.

\newsubsec{MTU Force}
Because the SE and CE are in series, we model their respective forces, $F_{se} \in \R$ and $F_{ce} \in \R$, as equal to the total force exerted by the MTU, denoted by $F_m \in \R$. Explicitly, we enforce $F_m = F_{se} = F_{ce}$. We independently model the individual element forces as depending on the previously defined force-length and force-velocity relationships:
\begin{align}
    & F_{ce}(l_{ce},v_{ce},s) = s \, F_{max} \, f_l(l_{ce}) \, f_v(v_{ce}), \\
    & F_{se}(l_{se}) = F_{max} \, f_{se}(l_{se}), 
\end{align}
where $s \in [0,1]$ is the activation level of the muscles, and $F_{max}\in\R$ is a constant parameter dictating the maximum allowable force of the MTU. Note that we assume muscle activation to be instantaneous.

\newsubsec{MTU Force-Torque Relationship}
The torque provided by the MTU, denoted by $u_m \in \R$, is calculated individually for each joint it spans using the following equations:
\begin{align}
    u_{m} &= r(\theta) \, F_m, \label{eq: musctorque} \\
    r(\theta) &= \begin{cases} r_0, &\text{for hip} \\
    r_0 \, cos(\theta - \theta_{max}),  &\text{ otherwise}
    \end{cases} \label{eq: leverarm}
\end{align}
where $r(\theta) \in \R$ is the length of the MTU lever-arm based on $r_0$ \new{(previously defined in Eq. \ref{eq: mtu_deltaL})}, and $\theta_{max} \in \R$ is the reference angle at maximum lever contribution. For MTUs that span two joints, the muscle torque of each joint is calculated using different muscle-specific maximum lever contribution reference angles $\theta_{max}$. \new{For details see \cite{geyer2010muscle}.}

\section{The Hybrid Zero Dynamics Method}
\label{sec: hzd}
Next, we present a high-level introduction of the HZD method (without the inclusion of muscle models) applied to the AMPRO3 prosthesis. For more details, we refer the reader to \cite{zhao2017multi}. Additionally, information on the mechanical design of AMPRO3 is outlined in \cite{zhao2017preliminary}.

 


\newsec{Human-Prosthesis Model for AMPRO3}
The human-prosthesis system is modeled as a seven-link \new{planar} model, illustrated in Fig. \ref{fig:model}c, with anthropomorphic parameters for the human segments (shown in blue), and parameters specific to the AMPRO3 prosthesis for the prosthetic segments (shown in red). Since the human user considered in this paper is not amputated, the model is asymmetric, with the knee of the prosthesis necessarily lower than the human knee.

The configuration space of the AMPRO3 prosthesis, assuming a floating-base convention \cite{grizzle2014models}, is defined as $\mathcal{Q} \subset \R^n$, where $n=9$ is the planar unconstrained degrees of freedom of AMPRO3. The base frame is defined as $q_B = (p, \theta_{\text{torso}}) \in SE(2)$ with $p \in \R^2$ and $\theta_{\text{torso}} \in SO(2)$ being the position and rotation of the floating base frame $R_B$ with respect to the world frame $R_w$.

We assume that the left leg is the intact leg and the right leg is the prosthetic leg. Hence, the human coordinates \new{$q_h = (\theta_{\torso},\theta_{\hip}^L,\theta_{\knee}^L,\theta_{\ankle}^L,\theta_{\hip}^R)^T$ consist of the torso angle and the joint angles of the human leg segments (left leg segments and right leg hip)}. The prosthetic coordinates $q_p = (\theta_{\knee}^R,\theta_{\ankle}^R)^T$ include the joint angles of the prosthetic segments. The generalized coordinate of the system is then defined as \new{$q = (p, q_h^T,q_p^T)^T$ }and the state space as $\mathcal{X} = T \mathcal{Q} \subset \R^{18}$ with coordinates $x =  (q^T,\dot{q}^T)^T$.


\newsec{Multi-Domain Hybrid System}
To capture the intrinsic nature of human walking, a multi-domain hybrid system is constructed for the human-prosthesis system model, with the goal of matching the temporal domain pattern observed in natural human walking \cite{ames2011human}. \new{Briefly, hybrid is used to refer to the involvement of both \textit{time-driven} and \textit{event-driven} events. Multicontact hybrid systems use multiple time-driven domains to describe different contact configurations. While such systems have been demonstrated to successfully yield multicontact locomotion \cite{zhao2017multi,zhao2017preliminary}, the inclusion of multiple domains significantly increases the complexity of the nonlinear optimization problem, making it more challenging to constrain the search space to achieve desire behaviors. In our work, we leverage muscle models to guide the optimization problem towards \textit{natural} multicontact walking gaits.}

As illustrated in Fig. \ref{fig:domains}, we construct a domain pattern with eight distinct domains (four in each step), and eight transitions between domains. Note that since our model is asymmetric, we need to consider an entire gait cycle from right heel strike to the next right heel strike, consisting of two individual steps. The domains within each step are named according to the contact points as: Double Support 2 (DS2$_{\{L,R\}}$), Double Support 3 (DS3$_{\{L,R\}}$), Single Support 2 (SS2$_{\{L,R\}}$), and Single Support 1 (SS1$_{\{L,R\}}$), where the subscript $\{L,R\}$ denotes either the left or right stance leg step. These domains are similar to the breakdown in \cite{reher2020algorithmic}.

Equipped with the domain definitions, we construct a directed cycle $\Gamma = (V,E)$ to describe our multi-domain hybrid system, with the vertices $V = \{v_1,\dots,v_8\}$ and edges $E = \{e_1,\dots,e_8\}$ illustrated in Fig. \ref{fig:domains}.
We denote the set of admissible domains by $\mathcal{D} = \{\mathcal{D}_v\}_{v \in V}$. The transitions between these domains are triggered by the set of guards, $S = \{S_e\}_{e \in E}$. The discrete dynamics of these transition events are denoted by $\Delta = \{\Delta_e\}_{e \in E}$. 

We can then formally define our full hybrid system as a tuple $\mathscr{HC} = (\Gamma,\mathcal{D},\mathcal{U},\mathcal{S},\Delta,FG)$,
where $\mathcal{U} = \{\mathcal{U}_v\}_{v \in V}$ is the set of admissible inputs and $FG = \{(f_v,g_v)\}_{v \in V}$ is the set of control systems with $(f_v,g_v)$ defining \new{the continuous dynamics $\dot{x} = f_v(x) + g_v(x) \, u_v$ for each domain with inputs $u_v = [u_{\hip}^L,u_{\knee}^L,u_{\hip}^L,u_{\ankle}^R,u_{\knee}^R,u_{\ankle}^R]^T$.} The continuous dynamics can be obtained using the Euler-Lagrangian equation as explained in \cite{zhao2017multi}. 

\newsec{Virtual Constraints}
The behavior of the hybrid system can be shaped using \textit{virtual constraints}, defined as the difference between the actual system outputs $y^a(q)$ and the desired outputs $y^d(q,\alpha)$. In our work, we describe the desired outputs using \new{B\'ezier polynomials with coefficients $\alpha$. To allow for discontinuities in the outputs between domains (necessitated by impact events), we describe the outputs using domain-specific B\'ezier polynomials with coefficients $\alpha_v$.}
For non-underactuated domains (DS2, DS3, SS2), a relative degree one output is explicitly included in the virtual constraints in order to regulate the forward progression of the system:
\begin{align}
    y_v(q,\alpha) = \begin{bmatrix}
    y_{1,v}(q,\dot{q},\alpha)\\
    y_{2,v}(q,\alpha)
    \end{bmatrix} = \begin{bmatrix}
    y_{1,v}^a (q,\dot{q}) - v_{\hip}\\
    y_{2,v}^a (q)-y_{2,v}^d(\tau(q),\alpha)
    \end{bmatrix},
\end{align}
Here $y_{1,v} \in \R$ denotes the domain-specific relative degree one output, defined as the difference between the actual hip velocity $ y_{1,v}^a (q,\dot{q})$ and the desired hip velocity $v_{\hip}$. The virtual constraints, $y_{2,v}(q,\alpha)$, denote the relative degree two output. Since the forward hip velocity is approximately constant during the progress of each step cycle, we define our phase variable \new{$\tau (q) = \frac{\delta_{p_{\hip}}(q) - \delta_{p_{\hip}}^+}{v_{\hip}}$,
where $\delta_{p_{\hip}}(q)$ is the linearized forward hip position} and $\delta_{p_{\hip}}^+$ is the hip position at the beginning of the step. We select the virtual constraints for each domain within one step to be the following: 
\begin{align*}
   y_{\text{DS2}} &= [v_{\hip},\theta_{\hip}^{\text{st}},\theta_{\hip}^{\text{sw}},\theta_{\knee}^{\text{sw}},\theta_{\text{ankle}}^{\text{sw}}]^T \\
   y_{\text{DS3}} &= [v_{\hip},\theta_{\hip}^{\text{st}},\theta_{\hip}^{\text{sw}},\theta_{\knee}^{\text{sw}}]^T \\
   y_{\text{SS2}} &= [v_{\hip},\theta_{\hip}^{\text{st}},\theta_{\knee}^{\text{st}},\theta_{\hip}^{\text{sw}},\theta_{\knee}^{\text{sw}},\theta_{\ankle}^{\text{sw}}]^T \\
   y_{\text{SS1}} &= [\theta_{\hip}^{\text{st}},\theta_{\knee}^{\text{st}},\theta_{\ankle}^{\text{st}},\theta_{\hip}^{\text{sw}},\theta_{\knee}^{\text{sw}},\theta_{\ankle}^{\text{sw}}]^T 
\end{align*}
where the superscripts (st, sw) denote either left ($L$) or right ($R$) for the stance and swing leg of the corresponding step. Note that the number of virtual constraints in each domain is dependent on the number of contact points.

\new{Lastly, the virtual constraints $y_v(q,\alpha)$ are driven to zero using a feedback linearizing controller $u^*(x)$, resulting in the closed loop dynamics $\dot{x} = f_{cl,v}(x) = f_v(x) + g(x)u^*(x)$.}

\newsec{Impact-Invariance Condition}
While the closed-loop dynamics of the designed trajectory may be stable, the system can destabilize at impact events. Thus, it remains to construct desired trajectories that are \textit{impact-invariant}. Since our hybrid system is a multi-domain system with both fully-actuated and under-actuated domains,  the entire system is impact invariant if the following individual \textit{impact-invariance conditions} are met for each transition:
\begin{align}
    \begin{cases}
    \Delta_e(S_e \cap \Z_{\alpha_v}) \subseteq \P\Z_{\alpha_v}, & e = \{3,7\},\\
    \Delta_e(S_e \cap \P\Z_{\alpha_v}) \subseteq \Z_{\alpha_v}, & e = \{4,8\},\\
    \Delta_e(S_e \cap \Z_{\alpha_v}) \subseteq \Z_{\alpha_v}, & \text{otherwise}.
    \end{cases} 
    \label{eq: impact-invariance}
\end{align}
Here we use $\P\Z_{\alpha_v}$ and $\Z_{\alpha_v}$ to denote the partial hybrid zero dynamics (PHZD) surface and HZD surface respectively:
\begin{align*}
    \Z_{\alpha_v} &= \{(q,\dot{q}) \in \mathcal{D}_v: y_{v}(q,\alpha_v) = 0, \, \dot{y}_{v}(q,\dot{q},\alpha_v) = 0\}, \\
    \P\Z_{\alpha_v} &= \{(q,\dot{q}) \in \mathcal{D}_v: y_{2,v}(q,\alpha_v) = 0, \, \dot{y}_{2,v}(q,\dot{q},\alpha_v) = 0\}.
\end{align*}
The system evolves on these surfaces when the virtual constraints are driven to zero 
. Note that $\P\Z_{\alpha_v}$ is a restriction of $\Z_{\alpha_v}$ for fully-actuated domains, in which the relative degree one output is used to regulate the system. In practice, impact-invariant periodic orbits are synthesized as solutions to a nonlinear optimization problem with constraints on the closed-loop dynamics and impact-invariance conditions.


 \section{Gait Generation with the Integrated Framework}
\label{sec: framework}

 
\new{Next, we present an integrated framework that enforces the various muscle-tendon unit properties introduced in Section \ref{sec: musclemodel} directly into the HZD gait generation framework introduced in Section \ref{sec: hzd}. First, we will present the details of the integrated framework. Then, we demonstrate its effect on the gait generation process by comparing gaits obtained with and without the inclusion of the musculoskeletal model.}

\newsec{Integrated Framework}
To generate stable impact-invariant periodic orbits, with the inclusion of the muscle models presented in Sec. \ref{sec: musclemodel}, we construct a nonlinear optimization problem of the form:
\par\vspace{-4mm}{\small
\begin{align*} 
   \{\bm{\alpha}^{*},{X}^{*}\} = \argmin_{\bm{\alpha},{X}} &~  \Phi_{\text{mCoT}}(X) \\
    \text{s.t.}\quad 
    & \textbf{C1.} \tag{Closed-loop Dynamics} \\
    & \textbf{C2.} \tag{Impact-Invariance Conditions} \\
    & \textbf{C3.} \tag{\new{Decision Variable Bounds}} \\
    & \textbf{C4.} \tag{Physical Constraints} \\
    & \textbf{C5-C12.} \tag{Muscle Model Constraints}
\end{align*}}
where $\bm{\alpha} = \{\alpha_v \mid v = 1,\dots,8\}$ is our collection of B\'ezier coefficients for each domain, and $X$ is the collection of all decision variables \new{$X = [X_{\text{NLP}},X_{\text{MUSC}}]^\top$  separated into the nominal variables, $X_{\text{NLP}}$, and the additional muscle model decision variables, $X_{\text{MUSC}}$. 
The nominal decision variables are constructed as $X_{\text{NLP}} = (x_0, \dots, x_N,T)$ with $x_i$ being the system state at the $i^{th}$ discretization for the duration $T$. The muscle model decision variables are similarly defined for the muscle states $x^{\text{musc}}$ as $X_{\text{MUSC}} = (x^{\text{musc}}_0, \dots, x^{\text{musc}}_N,T)$. Here, the muscle states include the MTU variables for each muscle $x^{\text{musc}} = \{ [ l_{ce}^{(i)}, l_{se}^{(i)}, F_{ce}^{(i)}, v_{ce}^{(i)}, s^{(i)} ]^\top \mid i = 1,\dots,10\}$. }

\begin{figure*}[tb]
    \centering
    \includegraphics[width=\linewidth]{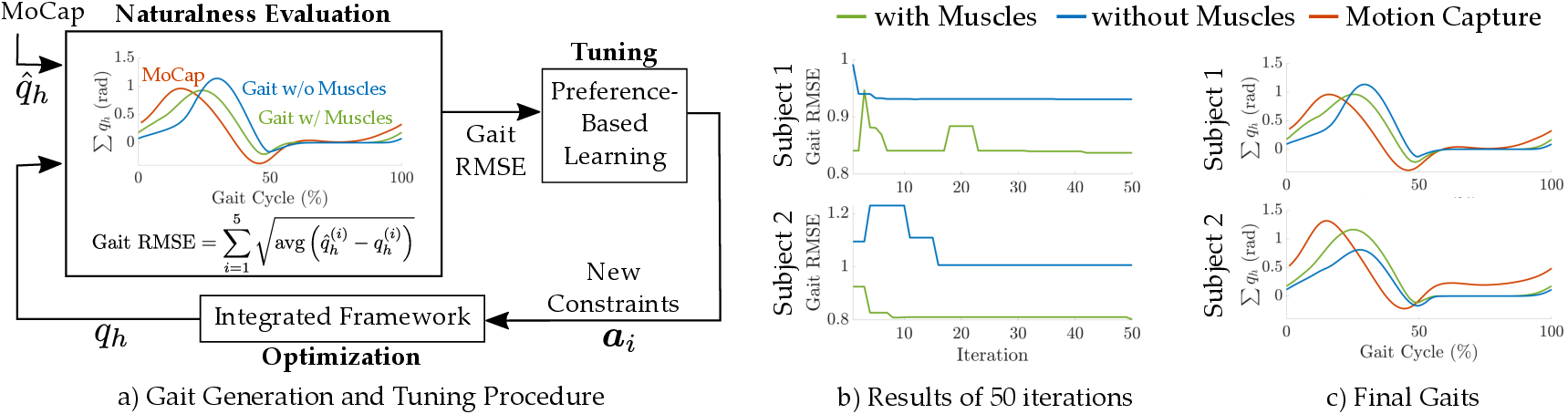}
    \caption{\new{Results of gait generated with and without the muscle models. a) Gait generation and tuning procedure. Note that the MoCap data are taken from \cite{camargo2021comprehensive} and matched to subjects by height and weight.  b) Gait RMSE of the optimal action identified by the algorithm at each iteration. c) The summed human joints angles of final gaits obtained after tuning.}}
    \label{fig:gait_rmse_plot}
\end{figure*} 

While the objective function can be arbitrarily defined, we intentionally select ours to be the mechanical cost of transport (mCoT), $\Phi_{\text{mCoT}} = \int \frac{P(t)}{mgv}dt$, since prior work has found it to yield natural and efficient locomotion \cite{hereid20163d}. 

The first four constraints (C1-C4) of our framework are standard to the HZD method\new{: C1 enforces the closed-loop dynamics of the system; C2 enforces the impact-invariance conditions described by Eq. \ref{eq: impact-invariance}; C3 constrains the decision variables as $X_{\text{min}}  \preceq X \preceq X_{\text{max}}$; and C4 enforces real world constraints such as contact constraints, as well as joint and torque limits.} The remaining constraints (C5-C12) are muscle model constraints, explicitly defined as:
\begin{ruledtable}
\vspace{-2mm}
{\textbf{\normalsize Muscle Model Constraints:}}
\par\vspace{-4mm}{\small 
\begin{align*} 
\textbf{C5.} &~\{F_m^{(i)} = F_{ce}^{(i)}(l_{ce}^{(i)},v_{ce}^{(i)},l_{se}^{(i)},s^{(i)}),  \,  \forall i = 1,\dots,10\}\\
\textbf{C6.} &~\{F_m^{(i)} = F_{se}(l_{se}^{(i)}),  \,   \forall i = 1,\dots,10\}\\
\textbf{C7.} &~ \{l_{ce}^{(i)} + l_{se}^{(i)} = l_{mtu}(q)^{(i)},  \,  \forall i = 1,\dots,10\} \\
\textbf{C8.} &~\{l_{ce}^{(i)} = \int v_{ce}^{(i)}dt,  \,  \forall i = 1,\dots,10\} \\
\textbf{C9.} &~ u_{\hip}^{L} = u_{m}^{(1h)} + u_{m}^{(2)} + u_{m}^{(3)}  \\
\textbf{C10.} &~ u_{\knee}^L = u_{m}^{(1k)} + u_{m}^{(4k)} - u_{m}^{(5)} \\
\textbf{C11.} &~ u_{\ankle}^L =u_{m}^{(4a)} + u_{m}^{(6)} - u_{m}^{(7)} \\
\textbf{C12.} &~ u_{\hip}^R = u_{m}^{8} + u_{m}^{9} - u_{m}^{10}
\end{align*}}\vspace{-6mm}\par
\end{ruledtable}
where $i = {1,\dots,10}$ denotes a specific muscle out of the ten muscles we consider, illustrated in Fig. \ref{fig:model}b. These muscles consist of seven muscles on the intact leg (hamstring (HAM), glutes (GLU), hip flexor (HFL), gastrocnemius (GAS), vastus (VAS), soleus (SOL),  tibialis anterior (TA)), and three muscles on the prosthetic leg (HAM, GLU, HFL).  

The first four muscular constraints (C5-C8) can be interpreted as \new{dynamic and kinematics} constraints acting on each MTU. The final four constraints (C9-C12) 
ensure that the actual human joint torque is equal to the sum of individual muscle torques. Depending on whether it is an extensor or flexor muscle, the torque is either applied towards the positive or negative direction. Note that since the HAM muscle span both the hip and knee joints, we use $u_m^{(1h)}$ and $u_m^{(1k)}$ to denote the torque HAM has on the these two joints respectively. Similarly, we use $u_m^{(4k)}$ and $u_m^{(4a)}$ to denote the knee and ankle joint torques resulting by GAS muscle. The explicit calculation can be found in Eq. \ref{eq: musctorque} with different reference angles in Eq. \ref{eq: leverarm}.


\newsec{Evaluation of the Integrated Framework}

\newsubsec{Optimization setup}
To evaluate our hypothesis that enforcing muscle model constraints would naturally lead to more anthropomorphic behavior, we \new{synthesized two variants of the optimization problem for comparison: 1) with muscles, which includes constraints C1-C12; and 2) without muscles, which only includes constraints C1-C4. In both variants, the optimization problem is constructed using FROST \cite{hereid2017frost}}. 

\new{We evaluated the naturalness of the gaits generated by the two variants via a custom metric defined as:
\begin{align}
    \text{Gait RMSE} = \sum_{i = 1}^5\sqrt{\text{avg} \left({\hat{q}_{h}^{(i)} - q_h^{(i)}} \right)}.
    \label{eq:rmse}
\end{align}
where $q_h^{(i)}$ denotes the angles of the $i^{th}$ joint of the human coordinates and $\hat{q}_{h}^{(i)}$ the corresponding joint angles recorded by MoCap. Specifically, the MoCap data used here are from  \cite{camargo2021comprehensive}  and  matched  to  subjects  by  height  and  weight. }



\begin{table}[b]
\centering
\new{
\caption{PBL Constraint Search Space}
\label{table:bounds}
\def\arraystretch{1.2} 
        \begin{tabular}{ | l | c | c|} 
        \hline
        Constraint Name & Constraint Values & lengthscales \\
        \hline
        $|\dot{x}| < a_1$ &  $a_1$:[15, 20] & 5 \\ 
        \hline
        $|\ddot{x}| < a_2$ & $a_2$:[70,80,90] & 10\\
        \hline
        $v_{\text{hip}} > a_3$ &  $a_3$:[0.3,0.4,0.5] (m/s) & 0.1 \\ 
        \hline
        $v_{\text{hip}} < a_4$ & $a_4$:[1.2,1.3,1.4] (m/s) & 0.1 \\ 
        \hline
        Min. Foot Clearance & $a_5$:[0, 0.013, 0.026, 0.039] (m) & 0.013 \\
        \hline
        $|\theta_{\text{torso}}| < a_6$ &  $a_6$:[0,0.1,0.2,0.3,0.4,0.5] (rad.)& 0.1 \\ 
        \hline
        $|\theta_{\text{hip}}| < a_7$ &  $a_7$:[20,35,50] (deg.)& 15 \\ 
        \hline
        $|\theta_{\text{ankle}}| < a_8$ &  $a_8$:[20,30,40] (deg.)& 10 \\ 
        \hline
    \end{tabular}}
\end{table}

 \begin{figure*}[tb]
     \centering
\includegraphics[width=\linewidth]{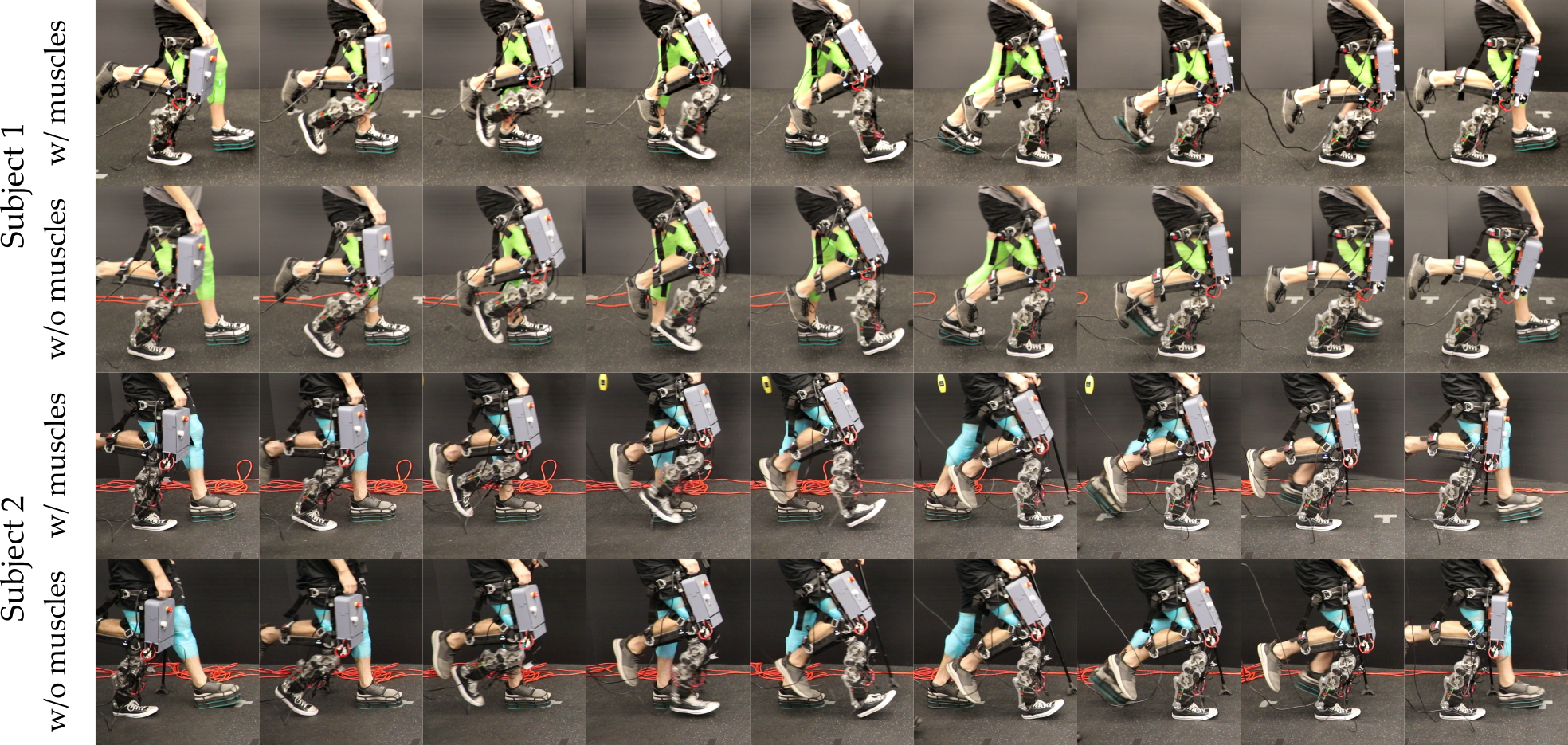}
     \caption{\new{Gait tiles of experimental demonstration on AMPRO3 for gaits generated without or with muscle muscle model for two subjects}}
     \label{fig:gaittiles}
 \end{figure*}
 
\newsubsec{Constraint Tuning via Preference-Based Learning} The bounds of C3 and C4 are commonly tuned in order to sufficiently constrain the optimization problem for convergence and to achieve desired behavior. 
Thus, to fairly compare gaits generated with and without the inclusion of the musculoskeletal model, we leverage preference-based learning to systematically identify the constraints that lead to the lowest Gait RMSE. The procedure of this framework is illustrated in Fig. \ref{fig:gait_rmse_plot}a. We specifically use the LineCoSpar \cite{tucker2020human} algorithm since it can navigate high-dimensional spaces and is robust to noisy feedback, but other Bayesian optimization techniques could also be used.


\new{In each iteration, we warm-start the optimization with the solution from the current best action according to the learning algorithm. To streamline the process, two types of feedback are automatically given to the algorithm. First, an ordinal label corresponding to either `converged' or `non-converged' is given based on the algorithm convergence status. Second, a pairwise preference is determined based on the Gait RMSE, where a lower RMSE gait would be preferred.  We construct the search space of the algorithm with the following dimensions as in Table \ref{table:bounds}.}

\newsubsec{Comparison of generated gaits}
\new{
This learning procedure was repeated for two subjects: subject 1 (Female, 172.7cm 65.7kg), subject 2 (Male, 180.3cm, 75kg). We plotted the Gait RMSE of gaits generated by the current best constraint parameters according to the algorithm at each iteration in Fig. \ref{fig:gait_rmse_plot}b. The inclusion of muscle models led to a smaller Gait RMSE compared with the ones generated by the non-muscle version throughout the tuning process (Fig. \ref{fig:gait_rmse_plot}b-c). This highlights the advantage of including muscle models in the gait generation, as it guided the optimization to find more natural solutions. }

 \begin{figure*}[tb]
    \centering
    \includegraphics[width=\linewidth]{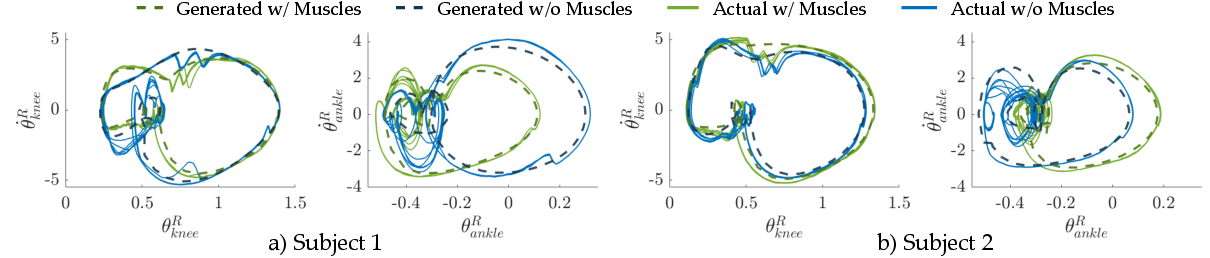}
    \caption{\new{Limit cycles illustrating the periodic stability achieved during experimental multicontact locomotion (10s of data plotted).}}
    \label{fig:limitcycle}
\end{figure*}

 \begin{figure*}[tb]
    \centering
    \includegraphics[width=\linewidth]{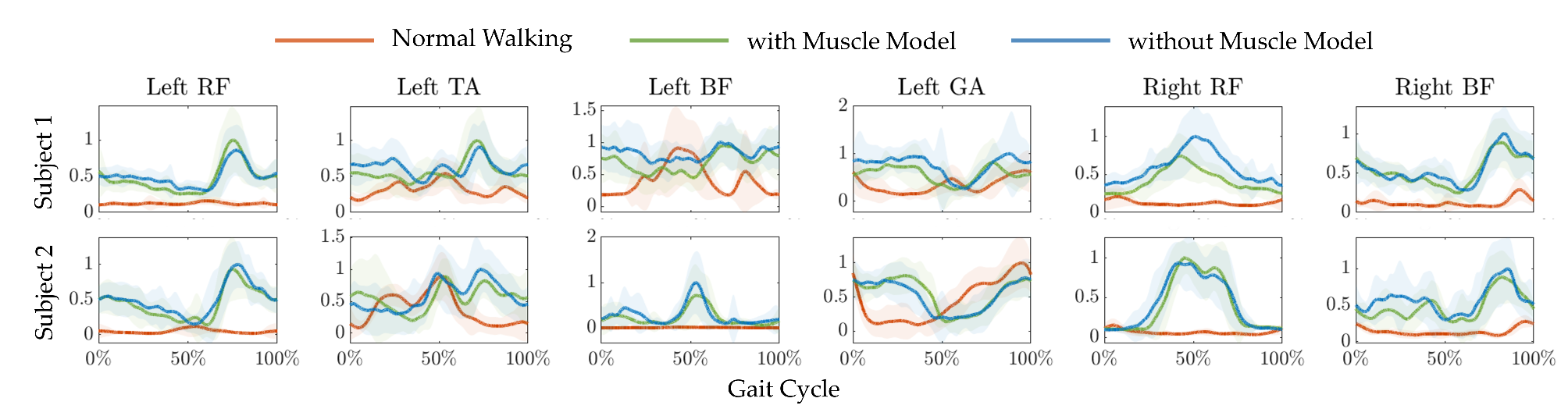}
    \caption{\new{EMG activity normalized over a full gait cycle for normal walking, prosthetic walking with gaits generated with or without the muscle model.}}
    \label{fig:emg_plot}
\end{figure*}

\section{Experimental Demonstration on AMPRO3}
\label{sec: expprocedure}
We experimentally deployed the two gaits obtained in the automated tuning procedure as having the lowest Gait RMSE (with and without the inclusion of muscle model constraints) on the dual-actuated transfemoral prosthesis, AMPRO3. This experiment was conducted for each of the two subjects, with the results highlighted in the supplemental video \cite{video}.

\newsubsec{Experiment Procedure} During the experiments, a non-disabled human user wore AMPRO3 using an adapter on the right leg (Fig. \ref{fig:exsetup}b). The joint-level trajectories of the gaits were tracked on the prosthesis with a PD controller. For an in-depth presentation of the hardware and control, see \cite{zhao2017multi}. 

First, the subject was asked to walk without the prosthesis over a self-selected speed, followed by walking with the prosthesis for the two prosthetic gaits. At the end of the testing, the subject was queried for a single pairwise preference. Note that the order of the gaits was randomized and the subject was not informed of the order. 
During all tests, electromyography (EMG) signals were recorded. Before recording, the subject was given enough time to adjust to the walking. In total, the activity of four muscles on the left leg, including rectus femoris (RF), tibialis anterior (TA), bicep femoris (BF), and gastrocnemius (GAS), and two muscles on the left leg (RF and BF) was recorded with the Trigno wireless biofeedback system (Delsys Inc.), as illustrated in Fig. \ref{fig:exsetup}c. 

\newsubsec{Experiment Results}
 A visualization of the experimental behaviors is provided in Fig. \ref{fig:gaittiles} via gait tiles spanning a complete gait cycle. \new{Both subjects strongly preferred the gait generated with the inclusion of the musculoskeletal model.}
 The stability of the executed gaits is portrayed in Fig. \ref{fig:limitcycle} by the periodicity of the limit cycles. It is important to note that achieving this experimentally stable multicontact locomotion is a direct result of leveraging the HZD method to formally generate impact-invariant output trajectories.



\new{The average EMG data over one gait cycle for each muscle after preprocessing is shown in Fig. \ref{fig:emg_plot}. We also calculated the RMSE between the EMG activity of the generated gaits and normal walking, defined as:
\begin{align}
    \text{EMG RMSE} = \sum_{i = 1}^6\sqrt{\text{avg} \left({\hat{s}_{\text{EMG}}^{(i)} - s^{(i)}_{\text{EMG}}} \right)},
    \label{eq:emg_rmse}
\end{align}
where $s_{\text{EMG}}^{(i)}$ denotes the muscle activation reflected by EMG signals for the $i^{th}$ muscle during the prosthetic walking and $\hat{s}_{\text{EMG}}^{(i)}$ denotes the corresponding muscle activation during normal unassisted walking.
The EMG RMSE are 1.58 and 1.84 for the gaits generated with muscles, and 1.80 and 2.34 for the gaits generated without muscles, for subject 1 and subject 2, respectively. The lower EMG RMSE suggests that the inclusion of the muscle model led to more \textit{natural} behavior. In addition, the inclusion of muscle model also results in less muscle activation on average.} Lastly, we observe that all prosthetic gaits yielded higher muscle activity than normal walking, which could be caused by factors such as the extra weight of the prosthesis or the misaligned knee joints.
However, when designing gaits for an amputee user, the human-prosthesis system would be more symmetric, which would likely to result in even more natural muscle activation.




\section{Conclusion}
\label{sec: conclusion}

This work demonstrates the first formal synthesis of stable multicontact locomotion using musculoskeletal models. Specifically, we directly enforce \new{muscle model constraints} in the HZD framework to experimentally realize both stable and natural robotic-assisted locomotion on the dual-actuated prosthesis AMPRO3 with two non-disabled users. We find that incorporating the muscle model guides the optimization problem towards uncovering periodic orbits that resemble natural bipedal locomotion. 

Our proposed framework is advantageous since it results in more natural behavior as compared to state-of-the-art. Additionally, it can be applied to a wide range of behaviors and/or robotic platforms, without relying on the availability of experimental data from human subjects or human-in-the-loop testing. \new{Lastly, even though the presented results are limited to planar locomotion, the framework can be extended to 3D locomotion by including muscles that act in the frontal plane (hip abductor and adductor \cite{song2013generalization})}.

\new{Since all physiological parameters (reference lengths, angle, etc.) were from \cite{geyer2010muscle}, which was intended for a non-disabled subject with different height and weight, it might be beneficial to calibrate these parameters of the muscle model to account for individual differences (especially for amputee users) and improve the prediction accuracy of the embedded muscle models, using methods similar to those in \cite{thatte2018method}.} Such prediction accuracy would further allow for targeted muscle behavior of the user for rehabilitation applications.


\bibliographystyle{IEEEtran}
\bibliography{references}

\end{document}